%% file: paper.tex
\documentclass[letterpaper, 10pt, conference]{ieeeconf}      

\IEEEoverridecommandlockouts                              

\usepackage{graphicx}
\usepackage{amsmath}
\usepackage{amssymb}
\usepackage{tabularx}
\usepackage{subcaption}
\usepackage{url}
\usepackage{cite}
\usepackage{algorithm}
\usepackage{algpseudocode}
\usepackage{blkarray}
\usepackage{prettyref}
\usepackage[colorlinks,bookmarksopen,bookmarksnumbered]{hyperref}
\usepackage{bbm}
\usepackage{dsfont}
\usepackage{multirow}
\usepackage{comment}
\usepackage{bm}
\usepackage{booktabs}
\usepackage{threeparttable}
\usepackage{soul}
\usepackage{xcolor}

\newrefformat{fig}{Fig.~\ref{#1}}
\newrefformat{sec}{Sec.~\ref{#1}}
\newrefformat{tab}{Tab.~\ref{#1}}
\newrefformat{alg}{Algo.~\ref{#1}}
\newrefformat{eq}{Eq.~\ref{#1}}

\title{\LARGE \bf Collaborative Assembly Policy Learning of a Sightless Robot} 

\author{Zeqing Zhang$^{1,*}$, Weifeng Lu$^{2,*}$, Lei Yang$^{1}$, Wei Jing$^{2}$, Bowei Tang$^{3}$, and Jia Pan$^{1,\dagger}$%
\thanks{$^*$ Equal Contributions. $^\dagger$ Corresponding Author. }
\thanks{This work was partially supported by NSFC-RGC Joint Research Scheme N$\_$HKU705/24, and the Natural Science Foundation of China (62461160309).}
\thanks{$^{1}$ Z. Zhang, L. Yang, J. Pan are with the School of Computing and Data Science, The University of Hong Kong, Hong Kong. {\tt\footnotesize zzqing@connect.hku.hk, \{lyang, jpan\}@cs.hku.hk}}
\thanks{$^{2}$ Weifeng Lu, Wei Jing are with NervAI, Hangzhou, China. {\tt\footnotesize \{luwf1992, 21wjing\}@gmail.com}}
\thanks{$^{3}$ Bowei Tang is with Shanghai Jiao Tong University, Shanghai, China. {\tt\footnotesize \ {aragorn.tang@gmail.com}}}
}

\begin{document}
\maketitle

\begin{abstract}
This paper explores a physical human-robot collaboration (pHRC) task involving the joint insertion of a board into a frame by a sightless robot and a human operator. 
While admittance control is commonly used in pHRC tasks, it can be challenging to measure the force/torque applied by the human for accurate human intent estimation, limiting the robot's ability to assist in the collaborative task.
Other methods that attempt to solve pHRC tasks using reinforcement learning (RL) are also unsuitable for the board-insertion task due to its safety constraint and sparse rewards. Therefore, we propose a novel RL approach that utilizes a human-designed admittance controller to facilitate more active robot behavior and reduce human effort.
Through simulation and real-world experiments, we demonstrate that our approach outperforms admittance control in terms of success rate and task completion time. Additionally, we observed a significant reduction in measured force/torque when using our proposed approach compared to admittance control.
The video of the experiments is available at \url{https://youtu.be/va07Gw6YIog}.
\end{abstract}


\input{intro}

\input{related}
\input{method}
\input{results}

\input{conclusion}

{
\bibliographystyle{IEEEtran}
\bibliography{references}
}

\end{document}

%% file: intro.tex
\section{Introduction}

\begin{figure}[!t]
\centering 
\includegraphics[width=0.48\textwidth]{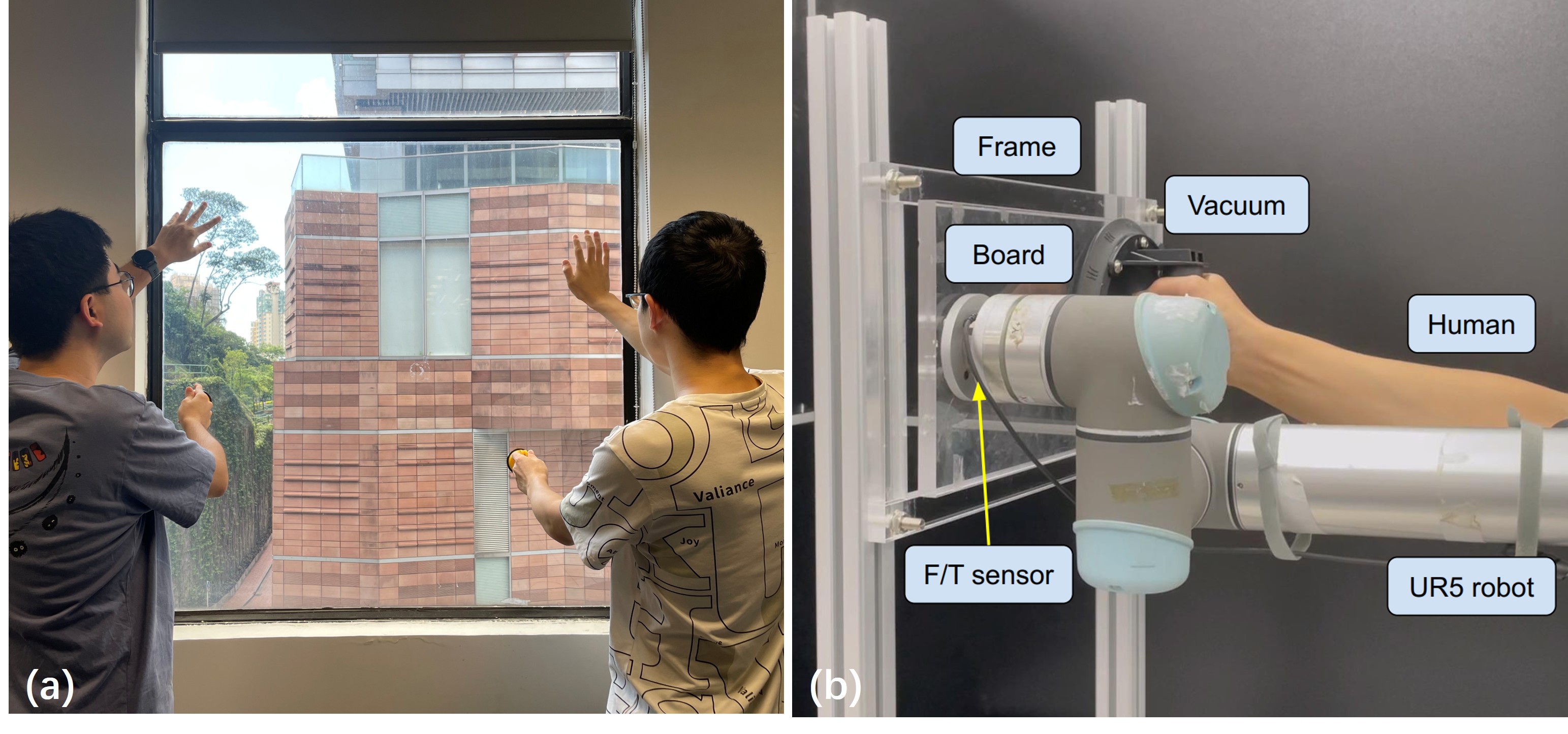}
\caption{(a) Installing a single pane of glass into a window frame by two people is a challenging task, even for skilled workers. (b) This paper presents a novel RL approach that employs a specialized admittance controller to facilitate human-robot collaboration for the board-insertion task, solely based on force feedback.}
\vspace{-6mm}
\label{fig:exp_scene} 
\end{figure}

Two or more humans handling heavy but fragile objects for accurate placement or assembly is a common occurrence in many daily and industrial domains, with the glazing task as a typical example (\prettyref{fig:exp_scene}-(a)), which involves installing glass in windows, doors, or other fixed openings. To reduce the need for human resources and effort, robots can be used as an alternative. However, due to the lack of perception accuracy, adaptive compliance, and planning intelligence, existing robots still struggle to accomplish these tasks independently. Thus, human-robot collaboration is considered a feasible solution, with successful examples like cooperative carrying \cite{gienger2018human} and co-manipulation for assembly \cite{cherubini2016collaborative}. 
During the collaboration, the human takes on the role of the leader while the robot acts as an assistant, carrying most of the load and understanding human intentions, such as identifying translations and rotations \cite{ansari2021task}, while being careful not to damage the object through the force it exerts.

This paper studies the \textit{human-in-the-loop board insertion task}, a simplified version of the challenging glazing task in a lab scenario. This task requires more precise position and force control than the general collaborative assembly tasks due to the \textit{millimeter tolerance} between the board and the frame, as shown in \prettyref{fig:exp_scene}-(b). Previous works typically use RGB-D cameras to provide visual information for collaborative assembly tasks, such as identifying the position and shape of a hole in the peg-in-hole task \cite{lee2020making} or estimating human intention in the collaborative carrying task \cite{yu2020human, xu2023visual}. A recent study \cite{mielke2017analysis} demonstrates that a team of two humans, with one blindfolded, can successfully perform a co-manipulation task, indicating that haptic rather than visual information is more crucial for communicating intent in co-manipulation. Inspired by this success, we consider the board-insertion task performed by a human-robot dyad that communicates through force sensing, with \textit{a sightless robot without vision sensors} serving as the replacement of the blindfolded participant.

Admittance control is a well-known solution for pHRC with continual contacts \cite{keemink2018admittance}. It generates compliant behavior by transferring force and torque to the desired movement using a second-order differential equation. However, the admittance control usually provides restricted assistance due to the lack of prior knowledge of human behavior patterns and task characteristics \cite{yu2020human}. This can lead to slow collaboration, particularly during the subtle alignment of the board and the frame, resulting in tedious human guidance.

In this paper, we employ reinforcement learning (RL) to teach a robot how to assist human operators in inserting a board into a frame. RL involves a trial-and-error process that enables a robotic agent to develop a control policy by exploring and interacting with the environment to attain a specific reward. Although it is straightforward to design a sparse reward for the board-insertion task according to whether the insertion is successful, creating a comprehensive dense reward for co-manipulation tasks in pHRC remains an open problem~\cite{mielke2017analysis}.

In an environment with sparse rewards, agents generally take longer to explore due to the lack of positive examples of rewarding actions. Safety concerns in our pHRC task impose additional constraints on agent exploration during training, making it more difficult to learn an effective control policy for assisting human operators. Inspired by residual RL, which combines a human-designed policy and a parametric policy to speed up the training process \cite{johannink2019residual}, we use a human-designed controller, specifically admittance control, to provide guidance for our RL policy learning. The key difference between our method and residual RL is that we use the human-designed controller to provide guidance at the early stage and gradually decrease its influence to learn a control policy at the end. Based on proximal policy optimization (PPO), we present an algorithm called policy-guided PPO (PGPPO). Real-world experiments verify that our PGPPO can achieve high-quality human-robot board insertion even with sparse rewards, outperforming admittance control.

\noindent{\textbf{Main contributions}}: 
\begin{itemize}
\item We develop an effective RL-based algorithm for human-robot co-manipulation that leverages admittance control as guidance to facilitate robot learning with a sparse reward. 

\item We evaluate the proposed approach in a \textit{millimeter-tolerance} board insertion task and explore the potential of using only haptic feedback for the human operator and a sightless robot. 

\item 
To the best of our knowledge (see \prettyref{tab:tasks}), this is the first attempt to investigate this pHRC task with millimeter tolerance between rigid bodies. 
\end{itemize}

The paper is organized as follows: \prettyref{sec:related} presents related work, \prettyref{sec:method} explains the problem formulation and solution details, \prettyref{sec:results} provides a discussion of results from simulations and experiments, and \prettyref{sec:conclusion} serves as the conclusion.

%% file: related.tex
\section{Related Work}
\label{sec:related} 

\subsection{Human-Robot Physical Collaboration}
\label{sec:HRPI}


Recent research shows that, with novel technologies, unexpected effects can also be achieved solely through haptic feedback \cite{zhang2025joint}.
By measuring the interaction force and torque of the human user, admittance control can be used to transfer haptic information to the desired robot movement \cite{keemink2018admittance}. 
This type of control is a mass-damper system, where the damping matrices largely affect human perception while the mass matrices are important for control stability \cite{lecours2012variable}. 
Robot behavior can be tuned more compliant/stiffer by decreasing/increasing the value of the damping. To achieve more flexible robot behavior, variable admittance control is used where the damping matrices are set manually, such as depending on the absolute value of the end-effector Cartesian velocity \cite{ficuciello2015variable}.

\begin{table}[ht]
    \centering
    \caption{Survey on Recent pRHC Tasks}
    \label{tab:tasks}
    \vspace{-5pt}
    \resizebox{0.48\textwidth}{!}{%
        \begin{tabular}{@{}ccccc@{}}
        \toprule
        \multirow{2}{*}{Related Work} & \multirow{2}{*}{Task Type} & Visual & F/T & Precision\\
        ~ & ~ & Feedback & Feedback & Tolerance\\
        \midrule\midrule
            F. Ficuciello \cite{ficuciello2015variable} & \multirow{2}{*}{2-DoF} & \multirow{3}{*}{No} & \multirow{3}{*}{Yes} & \multirow{3}{*}{N/A} \\
            S. Cremer \cite{cremer2019model} & \multirow{2}{*}{writing} & ~ & ~ & ~ \\
            J. R. Medina \cite{medina2019impedance} & ~ & ~ & ~ & ~ \\ \cmidrule(){1-5}
            \multirow{2}{*}{G. Kang \cite{kang2019variable}} & 2-DoF & \multirow{2}{*}{No} & \multirow{2}{*}{Yes} & \multirow{2}{*}{centimeter} \\ 
            ~ & tracking & ~ & ~ & ~ \\\cmidrule(){1-5}
            \multirow{2}{*}{X. Yu \cite{yu2020human}} & 2-DoF & \multirow{2}{*}{Yes} & \multirow{2}{*}{Yes} & \multirow{2}{*}{N/A} \\ 
            ~ & transporting & ~ & ~ & ~ \\\cmidrule(){1-5}
            \multirow{2}{*}{R. J. Ansari \cite{ansari2021task}} & 3-DoF & \multirow{2}{*}{No} & \multirow{2}{*}{Yes} & \multirow{2}{*}{N/A} \\
            ~ & handling & ~ & ~ & ~ \\\cmidrule(){1-5}
            E. A. Mielke \cite{mielke2017analysis} & 6-DoF & \multirow{2}{*}{Yes} & \multirow{2}{*}{Yes} & \multirow{2}{*}{N/A} \\
            W. Kim \cite{kim2017anticipatory} & manipulation & ~ & ~ & ~ \\ \cmidrule(){1-5}
            \multirow{2}{*}{Y. Yamakawa \cite{yamakawa2021development}} & 6-DoF & \multirow{2}{*}{Yes} & \multirow{2}{*}{Yes} & \multirow{2}{*}{millimeter} \\
            & peg-in-hole & ~ & ~ & ~ \\\cmidrule(){1-5}
            \multirow{2}{*}{\textbf{Ours}} & 6-DoF  & \multirow{2}{*}{No} & \multirow{2}{*}{Yes} & \multirow{2}{*}{millimeter}\\
            ~ & board insertion & ~ & ~ & ~ \\
        \bottomrule
        \end{tabular}
    }
    \vspace{-10pt}
\end{table}

To avoid a tedious and time-consuming human-engineered parameter tuning process, research has been conducted to find optimal damping matrices, such as using RL-based Fuzzy Q-Learning to regulate the damping matrices by minimizing jerk \cite{dimeas2015reinforcement, lu2020human}. Other related works have been proposed to minimize the cost energy of the motion by reducing the interaction force \cite{groten2012role}, the position error \cite{thobbi2011using}, or the task completion time \cite{duchaine2007general}. 
While these works can effectively determine the parameters of the dynamic systems of admittance control, they still require much effort to design a cost function for optimization purposes, and it is still unknown which objective(s) the approach should minimize to achieve the best performance in relevant physical human-robot collaboration tasks \cite{mielke2017analysis}. 
In this paper, we aim to propose an RL-based approach based on a binary reward (success/failure) to avoid the need for either human-engineered parameter tuning or cost function design for the co-manipulation task.

\subsection{Reinforcement Learning for Sparse Reward}
\label{sec:RL}

Reinforcement learning approaches face the challenge of sparse rewards due to the lack of positive data \cite{ladosz2022exploration}. To address this issue, researchers have proposed various solutions. For example, a specially designed reward function is proposed in the obstacle avoidance task \cite{han2022reinforcement}. Also, the curiosity about an agent can be used as an intrinsic reward signal for more intelligent exploration \cite{pathak2017curiosity}. 
Curriculum learning is another method that schedules the agent to solve a sequence of tasks with increasing complexity until the agent can solve the target task \cite{niu2023goats}.
Demonstration data is injected into the replay buffer to learn to perform long-horizon, multi-step robotics tasks successfully \cite{nair2018overcoming}.
Residual RL decomposes a control task into a structural part and a residual part and utilizes a conventional feedback controller and an RL controller to solve respective decomposed tasks~\cite{johannink2019residual}. However, the performance of residual RL relies on the conventional feedback controller, which can be challenging to design for tasks like co-manipulation.

Inspired by residual RL that leverages the conventional controller, we propose a policy-guided PPO that uses the admittance control as the initial policy for training the RL controller, addressing the challenge due to the sparse reward at the early exploration stage during training. As the training course proceeds, we gradually reduce the influence of the admittance control to derive an RL controller that can better collaborate with the human operator to perform the co-manipulation task. Experimental results show that our proposed method can learn a better control policy than the conventional admittance control.

%% file: method.tex
\section{Methodology}
\label{sec:method}

\subsection{Problem Formulation}
\label{sec:ac}

\begin{figure}[t]
\centering 
\includegraphics[trim=0 0 0 0, clip, width=0.98\linewidth]{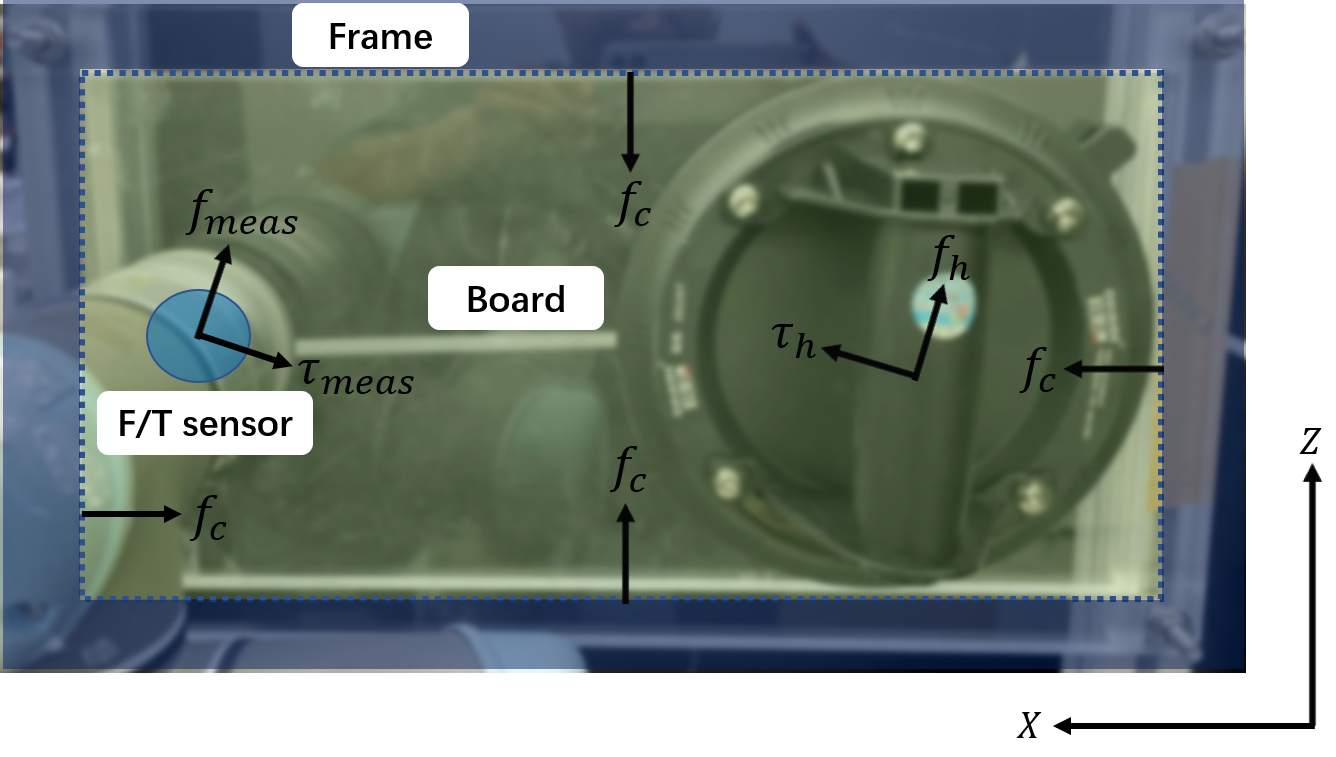}
\vspace{-5pt}
\caption{The front view when the board is inserted into the frame. $\bm{f}_c$ denotes the interaction force between the board and frame. $\bm{f}_h$ and $\bm{\tau}_h$ are the force and torque applied by the human. Force $\bm{f}_\text{meas}$ and torque $\bm{\tau}_\text{meas}$ are measured by the F/T sensor containing the coupled force/torque. Hence, the admittance control faces ambiguity in interpreting human intention. For example, when the human desires translation in the $Z$ direction by applying pure force in this direction, the torque in the $Y$ direction is generated and measured by the F/T sensor. Under the admittance control, the robot will simultaneously move along the $Z$-direction and rotate about the $Y$ axis, resulting in undesired assistance.}
\vspace{-5mm}
\label{fig:ft_analysis} 
\end{figure}

In this paper, we consider the task of board insertion into a rigid frame, which is a simplified version of the glazing task, performed by a dyad of a human operator and a sightless robot that communicates through force sensing. Although admittance control is often utilized to address co-manipulation tasks carried out by human-robot teams \cite{keemink2018admittance, lecours2012variable, li2018physical, ferraguti2019variable}, we note that using admittance control necessitates additional attention from human operators to control the applied force so that the robot can provide beneficial assistance (e.g., co-manipulation of the board in this scenario). The reason for this is that admittance control assumes human intention can be inferred from the measured forces that lead to the rigid motion of the target object. However, in the board insertion co-manipulation task, the force and torque applied by the human operator cannot be directly measured by the force/torque (F/T) sensor. For instance, when the operator applies a pure force (no torque) along one direction, the co-manipulation scenario depicted in \prettyref{fig:ft_analysis} can produce torque that can be perceived by the F/T sensor. This coupled F/T measurement may also result from interaction between the frame and the board. The resulting ambiguity has been explored in detail in recent work \cite{ansari2021task}. As a result, computing the force and torque applied by the operator would require extra information, such as the interacting location. While this may be feasible with sophisticated sensors, in this paper we consider a more general scenario where decoupled F/T measurement is not available.

To address this challenge, we propose an approach based on reinforcement learning that utilizes admittance control as prior knowledge to facilitate the training process and reduce the human operator's effort when performing this co-manipulation task.

Admittance control is formulated as follows:
\begin{equation}
\bm{M}_d \bm{\Ddot{x}}(t) + \bm{C}_d \bm{\Dot{x}}(t) + \bm{K}_d \bm{x}(t) = \bm{f}_{\text{meas}}(t),
\label{eq:ac_continuos} 
\end{equation}
where $\bm{M}_d$, $\bm{C}_d$ and $\bm{K}_d$ represent desired inertia, damping and spring matrices, $\bm{f}_{\text{meas}}(t)$ is the measured force (and torque), and $\bm{\Ddot{x}}(t)$, $\bm{\Dot{x}}(t)$ and $\bm{x}(t)$ denote Cartesian acceleration, velocity and position/orientation, respectively. 
In our RL formulation, the state of observation comprises the reference position/orientation and velocity at the previous time step, as well as the measured force/torque, denoted as $ \bm{s} = \left [\bm{x}_{r}(t), \bm{\Dot{x}}_{r}(t), \bm{f}_{\text{meas}}(t) \right]$. The action, denoted as $\bm{a} = \bm{\Dot{x}}_{r}(t+ \Delta{t})$, is the velocity command for the next step. This is comparable to the input/output of the admittance control problem.

The reward function is defined as
\begin{equation}
r(t) = \omega_1 \kappa  - \omega_2 * \frac{\left \| \bm{f}_{\text{meas}} \left( t \right) \right \|_2}{f_{\max}},
\end{equation}
where $\omega_1$ and $\omega_2$ are the hyperparameters that balance the two terms. 
The first term, $\kappa$, is a sparse reward that encourages the algorithm to accomplish the task without violating the safety constraint (explained below). It is defined as follows:
\begin{equation}
\kappa = 
\left\{
\begin{aligned}
& 200, && \text{task completed} \\ 
& -10, && \text{safety violation} \\ 
& 0. && \text{otherwise}
\end{aligned}
\right. \notag
\end{equation}
To prevent damage to the entire system (including both the robot and the frame) due to high measured force/torque, we introduce the safety constraint as the second term. The maximum force/torque value is denoted as $f_{\max}$. If the 2-norm value of measured force/torque exceeds $f_{\max}$, the task fails due to safety violations, and the process terminates immediately. We set $\omega_1=1$ and $\omega_2=0.02$ to prioritize the task completion over the safety. But once a policy that can achieve the board-insertion task with the human operator is learned, the second term weighted by $\omega_2$ will minimize the measured force/torque to ensure the safety constraint.

\subsection{Policy Guided PPO for Human-Robot Co-manipulation}
\label{sec:pgppo}

We introduce a policy-guided proximal policy optimization (PGPPO) algorithm that employs admittance control policy as guidance. Our approach draws inspiration from learning online with guidance offline (LOGO)~\cite{rengarajan2022reinforcement}, which considers two policy-updating steps: a policy improvement step and a policy guidance step. LOGO builds on the trust region policy optimization (TRPO) approach, and both policy improvement and guidance steps are constrained by the degree of similarity between the new and old policies, measured by Kullback–Leibler (KL) divergence, however, TRPO implementation is complex.
We present a novel algorithm to both simplify the formulation compared with LOGO~\cite{rengarajan2022reinforcement} and achieve better performance compared with standard PPO~\cite{schulman2017proximal}.

In our PGPPO algorithm, the \textit{policy improvement step} is the same as that in the standard PPO:
\begin{small}
\begin{equation}
\begin{aligned}
& g(\epsilon, A^{\pi_{\theta_k}}(\bm{s},\bm{a})) = \text{clip}\left(\frac{\pi_{\theta}(\bm{a}|\bm{s})}{\pi_{\theta_k}(\bm{a}|\bm{s})}, 1 - \epsilon, 1+\epsilon \right) A^{\pi_{\theta_k}}(\bm{s},\bm{a}), \\
& L(\bm{s},\bm{a},\theta_k,\theta) = \min\left(\frac{\pi_{\theta}(\bm{a}|\bm{s})}{\pi_{\theta_k}(\bm{a}|\bm{s})}  A^{\pi_{\theta_k}}(\bm{s},\bm{a}), g(\epsilon, A^{\pi_{\theta_k}}(\bm{s},\bm{a})) \right), \\
& \theta_{k+1/2} = \arg \max_{\theta} \underset{\bm{s},\bm{a} \sim \pi_{\theta_k}}{{\mathbb E}}\left[L(\bm{s}, \bm{a},\theta_k, \theta)\right],
\end{aligned}
\label{eq:improvement}
\end{equation}
\end{small}
where $\epsilon$ determines the allowable degree of deviation of the new policy from the old policy. During the \textit{policy guidance step}, our method employs the policy $\pi_H$ generated by admittance control, as described in \prettyref{eq:ac_continuos}, as guidance:
\begin{equation}
\begin{aligned}
& F(\bm{s},\theta_{k+1/2},\theta)=\min \left ( \frac{\pi_\theta \left ( \pi_H(\bm{s})|\bm{s} \right )}{\pi_{\theta_{k+1/2}} \left ( \pi_H(\bm{s})|\bm{s} \right )}, 1+\delta \right ), \\
& \theta_{k+1} = \arg \max_{\theta} \underset{\bm{s} \sim \pi_{\theta_k}}{{\mathbb E}}\left[F \left ( \bm{s},\theta_{k+1/2},\theta \right )\right].
\end{aligned}
\label{eq:guidance}
\end{equation}
This step facilitates learning by aligning the policy $\pi_\theta$ with the admittance control policy $\pi_H$. The hyperparameter $\delta$ determines the allowable degree of deviation of $\theta_{k+1}$ from $\theta_{k+1/2}$, similar to $\epsilon$.

In summary, PGPPO updates the policy by maximizing the expected cumulative reward and minimizing its similarity to the admittance control policy. As the admittance control policy generally yields sub-optimal results, we expect it to be more helpful during the early exploration stage and its influence to decrease as the training episode progresses. To achieve this, we only use the policy generated by admittance control at the initial training stage of the RL controller. We gradually reduce $\delta$ to a value close to zero as follows:
\begin{equation}
\label{eq:decay}
\delta_{k+1} \leftarrow \alpha \delta_{k}, \ \text{if} \ k > K,
\end{equation}
where $\alpha \in [0,1]$ is the decay coefficient, $k$ is the current training episode, and $\delta$ begins to decrease after the $K$-th episode. 
Moreover, since admittance control can gather state-action pairs $(\bm{s}_D,\bm{a}_D)$ in the board insertion task, we can utilize this demonstration data $\mathcal{D}$ to train PGPPO. The pseudo-code for our proposed PGPPO is given in~\prettyref{alg:pgppp}.

\subsection{Human Dynamics Model}
\label{sec:human_model}
Training an RL algorithm often requires a large number of training samples, which can be difficult to collect for human-in-the-loop tasks with sparse reward functions, such as our board-insertion task performed by a human-robot dyad. To address this, we propose using a human dynamic model to pre-train 
the PGPPO algorithm in a \textit{simulation environment}, thereby reducing the required number of real-world human demonstrations for training. The human dynamic model is based on the human limb dynamics and desired trajectories. The general human model introduced by~\cite{li2013human} is adopted, i.e.,
\begin{equation}
\label{eq:human_model}
-\bm{D}^k_h \Dot{\bm{x}}(t) + \bm{K}^k_h \left ( \bm{x}_d(t) - \bm{x}(t) \right) = \bm{f}_h(t),
\end{equation}
where $\bm{D}^k_h$ and $\bm{K}^k_h$ are the damping and stiffness matrices of a human limb at the $k$-th training episode; $\bm{x}_d(t)$ is the intended human motion trajectory; and $\bm{f}_h(t)$ is the force or torque exerted on the board by the human model.

To better model the variation among individual human operators, whenever the simulator is reset, we sample the damping ($\bm{D}^k_h$) and stiffness ($\bm{K}^k_h$) matrices from pre-defined, respective distributions listed in \prettyref{tab:parameter}, called \textit{domain randomization} \cite{tobin2017domain}. Hence, in each episode, the human dynamic model may vary to mimic the individual differences among human operators.

To model an intended motion trajectory $\bm{x}_d(t)$ of the human operator, we adopt the approach in~\cite{medina2019impedance}, which approximates the trajectory as a cubic spline interpolation:
\begin{equation}
\label{eq:human_intention}
\bm{x}_d(t) = 
\left\{
\begin{aligned}
& \bm{a}t^3 + \bm{b}t^2 + \bm{c}t + \bm{d}, \ t \leqslant T \\ 
& \bm{x}_f, \ t > T. \\ 
\end{aligned}
\right. \notag
\end{equation}
where $t$ and $T$ are the current time and total planning time, and $\bm{x}_f$ is the board's target position and orientation when it is successfully inserted into the frame. The parameters $\bm{a},\bm{b},\bm{c},\bm{d}$ are calculated as
\begin{equation}
\begin{aligned}
& \bm{c} = \bm{v}_i, && \bm{d} = \bm{x}_i, \\
& \bm{a} = \frac{2(\bm{d}-\bm{x}_f)+(\bm{c}+\bm{v}_f)T}{T^3}, && \bm{b} = \frac{\bm{v}_f-\bm{c}-3\bm{a}T^2}{2T},
\end{aligned} \notag
\end{equation}
where $\bm{x}_i$ is the board's initial position and orientation, and $\bm{v}_i$ and $\bm{v}_f$ are the board's initial and final velocities, which include both translational and angular components.
Our experiment demonstrates that this simplified human dynamics model can help the robot learn a workable policy in simulation, allowing it to collaborate with human operators in real life to successfully complete the board insertion task.

\begin{algorithm}[!tb]
\caption{Policy guidance proximal policy optimization}\label{alg:pgppp}
\begin{algorithmic}[1]
\State Input: Admittance controller $\pi_H$ (\prettyref{eq:ac_continuos}) and/or demonstration data $\mathcal{D}$, initial policy parameters $\theta_0$, initial value function parameter $\phi_0$.
\For{$k=0,1,2,\cdots$}
\State Collect set of trajectories $\mathcal{D}_k={\tau_i}$ by running policy
\Statex \ \ \ \ $\pi_k=\pi(\theta_k)$ in the environment.
\State \textbf{\textit{Policy improvement step}}: \prettyref{eq:improvement}
\State \textbf{\textit{Policy guidance step}}:
\If{only $\pi_H$ is known}
\State \prettyref{eq:guidance}
\normalsize \ElsIf {only $\mathcal{D}$ is known}
\State $G(\bm{s},\theta_{k+1/2},\theta)=\min \left ( \frac{\pi_\theta \left ( \bm{a}_{D}|\bm{s}_{D} \right )}{\pi_{\theta_{k+1/2}} \left ( \bm{a}_{D}|\bm{s}_{D} \right )}, 1+\delta \right )$
\State $\theta_{k+1} = \arg \max_{\theta} {{\mathbb E}}\left[G\left ( \bm{s},\theta_{k+1/2},\theta \right )\right]$
\ElsIf{both $\pi_H$ and $\mathcal{D}$ are known}
\State $\theta_{k+1} = \arg \max_{\theta} {{\mathbb E}}\left[F + G\right]$
\normalsize \EndIf
\State Fit value function by regression on mean-squared
\Statex \ \ \ \ error:
\State $\phi_{k+1}=\text{arg} \ \underset{\phi}{\min} \frac{1}{|D_k|T} \sum_{\tau \in D_k} \sum_{t=0}^T ( V_\phi ( \bm{s}_t )-\hat{R}_t )^2$
\State Decay $\delta$ by~\prettyref{eq:decay}.
\EndFor
\end{algorithmic}
\end{algorithm}

%% file: results.tex
\section{Experiment Setup and Results}
\label{sec:results}

Our method is trained in a simulation environment, and we evaluate its performance in both simulated and real-world setups. Neither the human dynamics nor the task characteristics, such as the board's target position, are known to the agent in these setups. We expect the agent to learn this information through interaction with the human operator and the environment.

\subsection{Simulation Setup}
\label{sec:sim_setup}

We use PyBullet as our physics simulator for robot learning, as it is fast and user-friendly. Our simulation environment includes a UR5 robot, an F/T sensor mounted on the robot's end-effector, a board, and a frame.

To minimize the sim-to-real gap, we reference the real-world experimental setup to set the simulation parameters listed in \prettyref{tab:parameter}. Parameters that can be directly measured in the real world, such as the position and orientation of the frame, the board-frame tolerance, the board's mass, and the noise level of the F/T sensor, are set to their exact values. However, other parameters, such as the stiffness and damping coefficients, are either challenging to measure or subject to change over time. To address this, we utilize the uniform domain randomization method to sample a range of simulated environments with randomized properties, including the human stiffness and damping values mentioned in \prettyref{sec:human_model} and the stiffness of the board and the frame.

\subsection{Simulation Results}
\label{sec:sim_result}
We test three types of methods, namely 1) Admittance Control (AC), 2) standard PPO, and 3) PGPPO with different types of prior knowledge as guidance in the simulation. Here is a list of methods that we compare:
\begin{itemize}
\item Admittance control (AC, \prettyref{eq:ac_continuos});
\item Standard PPO without any guidance knowledge;
\item PGPPO with the admittance control $\pi_H$ (\prettyref{eq:ac_continuos});
\item PGPPO with real-world human demonstration data $\mathcal{D}$. 
\item PGPPO with both $\pi_H$ and $\mathcal{D}$;
\end{itemize}

While we train the PGPPO controller in a simulated environment, the human demonstration data $\mathcal{D}$ were collected in the real-world setting (see \prettyref{fig:exp_scene}-(b)) by asking a human operator to work with AC to perform the board insertion task. Only success cases are included to guarantee the quality of demonstration data. To ensure fairness in comparison, we employ identical admittance control as guided policy $\pi_H$ and generate demonstration data $\mathcal{D}$ for all three variations of PGPPO.

\begin{figure}[t]
\centering 
\includegraphics[trim=30 20 50 20, clip, width=0.95\linewidth]{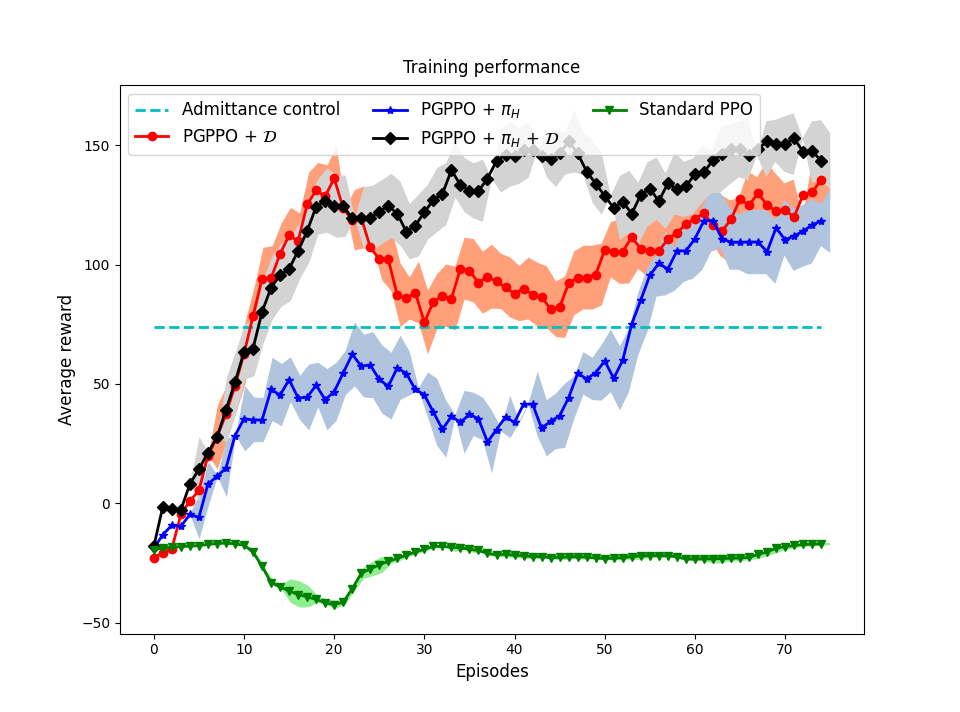}
\vspace{-5pt}
\caption{Learning curves of different methods in simulation. All PGPPOs with different types of prior knowledge achieve better performance than admittance control. Standard PPO cannot learn a policy to finish the insertion task.}
\label{fig:training_result} 
\end{figure}

The training performance over 75 episodes is displayed in \prettyref{fig:training_result}. Our observations indicate that \textit{Standard PPO} is not suitable for the board-insertion task performed by a human-robot dyad due to the sparsity of rewards. \textit{AC} outperforms \textit{Standard PPO}, demonstrating its effectiveness as a guidance method for training our proposed RL algorithm.
We would like to highlight that all PGPPO variants, which incorporate different forms of prior knowledge as guidance, outperform both \textit{AC} and \textit{Standard PPO}. This validates the effectiveness of our algorithmic design.

We conducted an ablation study to investigate how different forms of guidance can enhance our proposed PGPPO method. Our findings show that \textit{PGPPO with $\mathcal{D}$} learns effectively when trained using demonstration data containing only successful samples. However, its performance may fluctuate over the training course.
Another variant, \textit{PGPPO with $\pi_H$}, employs the admittance control policy as guidance. Although it exhibits slower learning efficiency and only outperforms \textit{AC} in the later training stage, it eventually converges to a performance level similar to \textit{PGPPO with $\mathcal{D}$}.
The third variant, \textit{PGPPO with both $\pi_H$ and $\mathcal{D}$}, synergistically leverages both the human demonstration data and the admittance control policy. This variant achieves the same level of learning efficiency as \textit{PGPPO with $\mathcal{D}$} and attains the highest reward among the three variants. From these results, we conclude that using the AC policy as guidance is beneficial for training our RL controller at the early stage while human demonstrations can provide substantial positive examples for stabilizing our RL controller's performance during the course of training.

\prettyref{tab:sim_rl_ac} presents the performance of the PGPPO variants and the admittance control in terms of the average success rate and completion time of the board-insertion task over 25 trials. A trial is considered a failure if the F/T sensor detects force or torque values exceeding the prescribed thresholds (i.e., violating the safety constraint). The PGPPO variants achieve a higher success rate and complete the task in a shorter time than VAC. Notably, the standard deviation of the task completion time for \textit{PGPPO with both $\pi_H$ and $\mathcal{D}$} is significantly smaller than that of VAC and the other PGPPO variants.
Based on the above results, we have found that using \textit{PGPPO with both $\pi_H$ and $\mathcal{D}$} is more advantageous compared to other PGPPO variants. Therefore, we use \textit{PGPPO with both $\pi_H$ and $\mathcal{D}$} in our real-world experiments.

\begin{table}
    \caption{Parameters of the simulation setup. The lower and upper bounds of $\bm{D}_h$ and $\bm{K}_h$ are reported, respectively. ($U$: Uniform distribution. $\mathcal{N}$: Normal distribution.)}
    \vspace{-5pt}
    \centering    
    \resizebox{0.43\textwidth}{!}{%
    \begin{tabular}{@{}cc c @{}} 
    \toprule
        Parameter &  Value or Range & Unit\\ \midrule \midrule
        \multirow{2}{*}{$\bm{D}_h$}     &  $\text{diag} \left ( \left [ 5,5,5,0.05,0.05,0.05 \right ] \right )$ & \multirow{2}{*}{$kg/s$} \\ 
        ~    &  $\text{diag} \left ( \left [ 375,375,375,2,2,2 \right ] \right )$ &  ~\\ \midrule 
        \multirow{2}{*}{$\bm{K}_h$}    &   $\text{diag} \left ( \left [ 200,200,200,2,2,2 \right ] \right )$& \multirow{2}{*}{$N/m$}\\ 
        ~    & $\text{diag} \left ( \left [ 1500,1500,1500,10,10,10 \right ] \right )$  &  ~\\ \midrule
        $\bm{M}_d$    &  $\text{diag} \left ( \left [ 0.5,0.5,0.5,0.1,0.1,0.1 \right ] \right )$ & $kg$ ~\\ \midrule
        $\bm{C}_d$    &  $\text{diag} \left ( \left [ 12.5,12.5,12.5,1.5,1.5,1.5 \right ] \right )$ & $kg/s$ ~\\ \midrule
        $\bm{K}_d$    &  $\text{diag} \left ( \left [ 1.5,1.5,1.5,4.5,4.5,4.5 \right ] \right )$  & $N/m$ ~\\ \midrule
        Board stiffness    &   $U(10^5, 1.5 \times 10^5)$ & $N/m$\\ \midrule
        Frame stiffness    &   $U(10^5, 1.5 \times 10^5)$ & $N/m$\\ \midrule
        $F \ \text{noise}$ & $\mathcal{N}(0 ,1/16)$ & $N$\\ \midrule
        $T \ \text{noise}$ & $\mathcal{N}(0, 1/750)$& $Nm$\\ \midrule
        Board size    &  $0.4 \times 0.2 \times 0.015$ & $m$\\ \midrule
        Board mass    & $0.714$& $kg$\\ \midrule
        Vacuum mass   & $0.418$ & $kg$\\ 
    \bottomrule
    \end{tabular}
    }
    \vspace{-15pt}
    \label{tab:parameter}
\end{table}

\begin{table}[!ht]    
    \caption{Comparison of the performance of PGPPOs with different prior knowledge and admittance control in simulation. (S.R.: success rate, Time: completion time. F and T in the Cause of Failure: force and torque.)}
    \vspace{-5pt}
    \centering    
    \resizebox{0.46\textwidth}{!}{%
    \begin{tabular}{@{}cccc@{}} 
    \toprule
        Method & S.R. & Time (s) & Cause of Failure   \\ \midrule\midrule 
        PGPPO with $\pi_H$ & $60\%$ & $14.21 \pm 2.60$ & F: $30\%$, T: $70\%$ \\ \midrule
        PGPPO with $\mathcal{D}$ & $64\%$ & $13.17 \pm 2.45$ & F: $37.5\%$, T: $62.5\%$ \\ \midrule 
        PGPPO with $\pi_H$ and $\mathcal{D}$ & $84\%$ & $12.36 \pm 1.78$ & F: $25\%$, T: $75\%$ \\ \midrule 
        Admittance control & $56\%$ & $14.52 \pm 2.77$ & F: $22.2\%$, T: $77.8\%$ \\
    \bottomrule
    \end{tabular}
    }
    \label{tab:sim_rl_ac}
    \vspace{-10pt}
\end{table}

\subsection{Real-World Experiment Setup}
\label{sec:exp_setup}

The real-world experiment is set up as shown in \prettyref{fig:exp_scene}-(b) and \prettyref{fig:ft_analysis}, utilizing a UR5 robot, an ATI Mini45 F/T sensor installed at the robot's end-effector, a frame, and an acrylic board to be manipulated. The human operator uses a vacuum to manipulate the board, and the frame is positioned directly in front of them. Initially, the board and frame are parallel in the $XZ$-plane. For this experiment, the action space is simplified to four dimensions: translation along $X, Y$, and $Z$, and rotation about $Y$. The parameters for the board's mass and the vacuum device are listed in \prettyref{tab:parameter}.

The human operator is responsible for applying force/torque on the vacuum device to insert the board into the frame alongside the robot. As mentioned in the simulation setup, the insertion task will fail if the safety constraint is violated. To safeguard the UR5 manipulator, the force and torque thresholds in the safety constraint are set at $80N$ and $8Nm$, respectively. If these thresholds are exceeded, an emergency stop will be triggered.

\begin{figure*}[!tbp]
\centering
\includegraphics[width=0.98\textwidth]{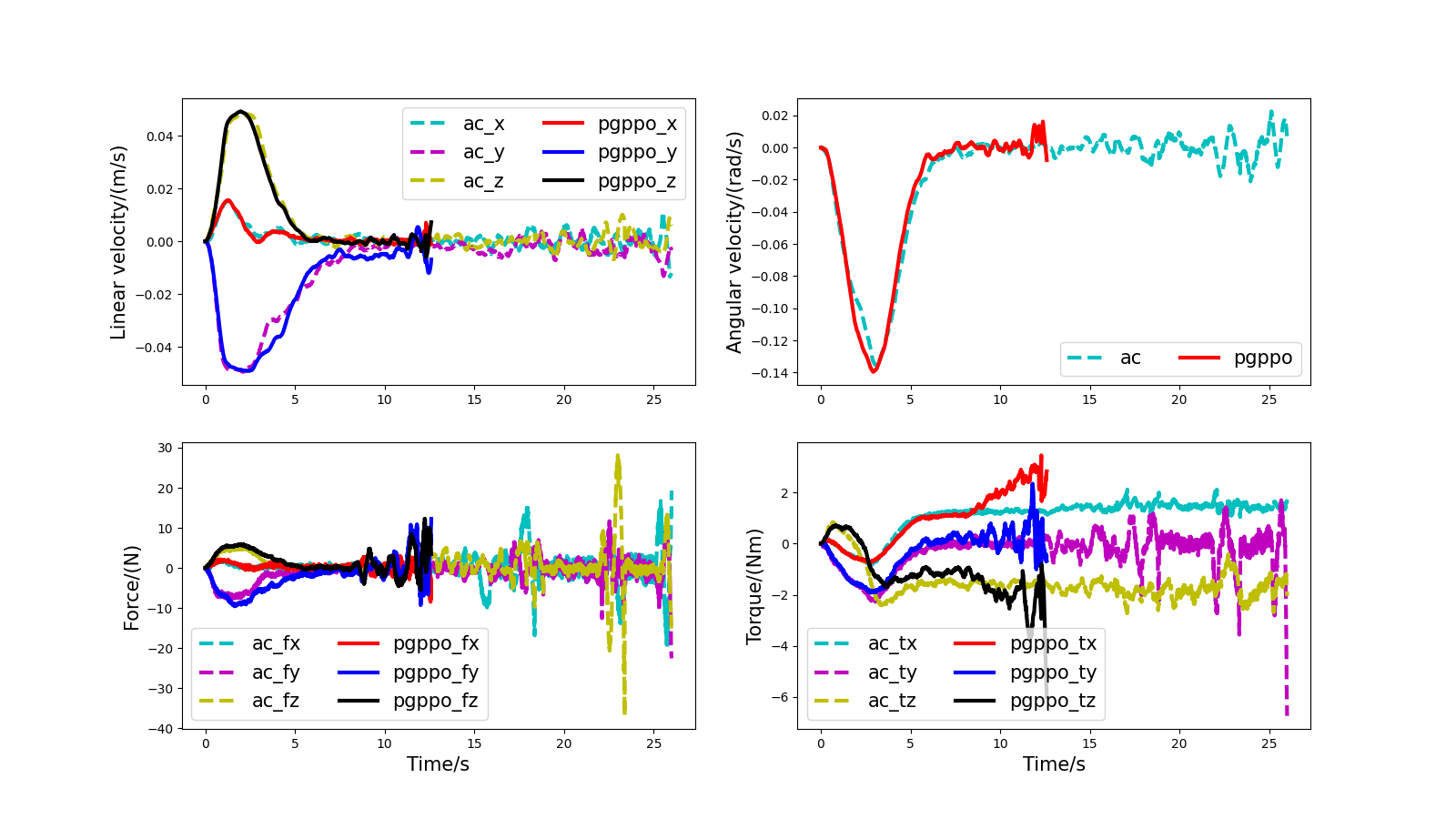}
\vspace{-20pt}
\caption{The robot end-effector velocity (upper row) and F/T sensor data (bottom) in real-world experiments. There is little difference between PGPPO and admittance control in the approaching phase. But time spent in the inserting phase using PGPPO is much shorter.}
\label{fig:exp_ac_rl}
\end{figure*}

\begin{figure*}[!tbp]
    \centering
    \includegraphics[width=0.98\textwidth]{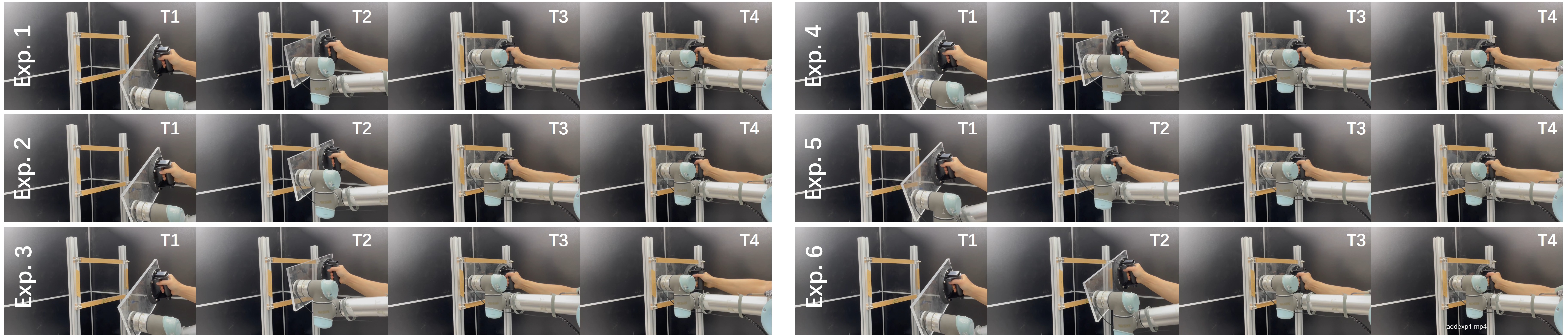}
    \caption{The process of the board insertion tasks. T1, T2, T3, and T4 are the initial, approaching, inserting, and completed states, respectively. See video for more experiments.}
    \label{fig:exp_phase}
    \vspace{-10pt}
\end{figure*}

\begin{table}[t]
    \caption{Real-world experiment results.}
    \vspace{-5pt}
    \centering    
    \resizebox{0.46\textwidth}{!}{%
    \begin{tabular}{@{}cccc@{}} 
    \toprule
        Method & S.R. & Time (s) & Cause of Failure   \\ \midrule\midrule 
        PGPPO (\textbf{Ours}) & $80\%$ & $10.23 \pm 1.47$ & F: $33\%$, T: $67\%$ \\ \midrule
        Admittance control & $60\%$ & $13.31 \pm 2.87$ & F: $24\%$, T: $76\%$ \\
    \bottomrule
    \end{tabular}
    }
    \label{tab:exp_rl_ac}
    \vspace{-15pt}
\end{table}

\subsection{Real-World Experiment Results}
\label{sec:exp_result}

\begin{figure}[t]
    \centering
    \includegraphics[trim=20 20 10 20, clip, width=0.8\linewidth]{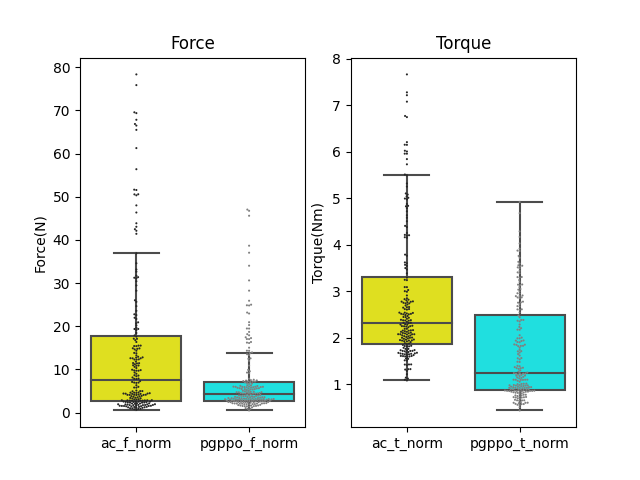} 
    \label{fig:f_ac_rl}
    \caption{Points in the boxplots show the L2-norm of measured force/torque in the inserting phase. 
    The median values of F/T using PGPPO are smaller than those using admittance control. 
    }
    \label{fig:exp_ft}
    \vspace{-5mm}
\end{figure}

In the real-world experiments, we tested two methods: 1) \textit{AC} and 2) \textit{PGPPO with both $\pi_H$ and $\mathcal{D}$}, which performs the best among the three variants as shown in~\prettyref{tab:sim_rl_ac}. 
In this real-world experiment, our method \textit{PGPPO with both $\pi_H$ and $\mathcal{D}$}, trained on a collection $\mathcal{D}$ of human demonstration data, is evaluated; no participant's data were used for fine-tuning the learned control policy of our PGPPO.

Five volunteers (3 males and 2 females) participated in the experiment, with an experimenter providing an introduction to the process. Participants were asked to test both methods, including the proposed PGPPO and the admittance controller, without prior knowledge of our hypothesis. To prepare for the formal experiments, participants were allowed to practice the insertion task with the robot several times. Throughout both the practice and formal experiments, participants were advised to be patient, as the tolerance between the board and frame was at the millimeter level. They were warned that excessive force or torque could trigger an emergency stop, causing the robot to fail to complete the task.

Each method was tested 30 times with 5 participants, and the success rate and mean completion time are reported in \prettyref{tab:exp_rl_ac}. Compared to admittance control, PGPPO significantly improved the success rate of the insertion task from 60\% to 80\%. Among successful cases, PGPPO had a shorter completion time with a smaller standard deviation, consistent with our simulation results. In failed cases of PGPPO, we observed a higher percentage of failures caused by torque exceeding the safety threshold than that caused by force exceeding the threshold.

In addition, we plot the averaged robot Cartesian velocity and F/T sensor data for all trials of both methods in \prettyref{fig:exp_ac_rl}. The co-manipulation process can be divided into two phases: the approaching phase and the inserting phase. These phases are defined based on the time of the first contact between the board and frame, as shown in \prettyref{fig:exp_phase}. In the approaching phase, the operator holds the board while approaching the frame, corresponding to the smoother part of the plot curves. In the inserting phase, fine-grained manipulation occurs as the operator carefully inserts the board into the frame.
The approaching phase duration shows little difference between the two methods, about $6.73$s for admittance control and $6.2s$ for PGPPO. However, the inserting phase duration using PGPPO is much shorter than admittance control ($8.07 \pm 2.13s$ vs. $12.07 \pm 1.27s$), demonstrating the effectiveness of the PGPPO approach in a real-world setting.

To achieve fine-grained co-manipulation and ensure safety by avoiding large force/torque during the task, it is desirable that the F/T readings measured by the sensor remain small. This indicates a consensus between the human operator and the robot, resulting in a smooth collaboration between the human-robot team.
We examined the force and torque distributions measured with PGPPO and admittance control in \prettyref{fig:exp_ft}. Each boxplot in this figure shows the 2-norm value of the force and torque. Differences can be observed between the force/torque distributions of the two methods. The median values and outliers of the measured force/torque with the PGPPO approach are smaller than those using admittance control, which can explain the higher success rate achieved with PGPPO.


Finally, we test whether the difference between PGPPO and admittance control in terms of the measured force and torque is statistically significant using the Mann-Whitney U test\cite{hollander1999solution}, a nonparametric test that does not assume the data follows a specific distribution. 
The null hypothesis $H_0$ is that the 2-norm of the instantaneous force or torque using PGPPO is statistically greater than or equal to that using admittance control, denoted by $H_0: FT_\text{pgppo} \geq FT_\text{ac}$. The alternative hypothesis $H_1$ is denoted by $H_1: FT_\text{pgppo} < FT_\text{ac}$. To reject the null hypothesis $H_0$ in favor of the alternative hypothesis $H_1$, a confidence level of 95\% is required. 
We uniformly sampled the force/torque signals read by the F/T sensor and calculated the $p$ value based on these sampled points.
The $p$-values corresponding to force and torque are $0.027$ and $0.043$, respectively, which are smaller than 0.05. Therefore, we can reject the null hypothesis in favor of the alternative: $H_1: FT_\text{pgppo} < FT_\text{ac}$. In summary, we conclude that the 2-norm value of F/T data using PGPPO is significantly smaller than that using admittance control in a statistical sense. Thus, PGPPO has the ability to decrease the value of contact force and torque and improve the success rate of the insertion task.

%% file: conclusion.tex
\section{Conclusion and Future Work}
\label{sec:conclusion} 

In this paper, we investigate the physical human-robot collaboration task of inserting a board into a frame performed by a human operator and a sightless robot. Due to the lack of a vision sensor, the human-robot team can only communicate with each other through force feedback. Due to the binary reward of this task, we present a method, Policy-Guided PPO, that utilizes the admittance controller and demonstration data to facilitate policy learning.
We validated our design choices through simulation and real-world experiments. In the simulation, we demonstrated that incorporating both admittance control policy and demonstration data leads to a fast convergence rate and stable performance during training. In the real-world setup, we compared the proposed PGPPO to the admittance controller. The results show that the PGPPO policy achieved a higher success rate ($80\%$) and shorter task completion time ($\sim 10s$) for the human-robot team. Additionally, we observed that the measured force/torque in PGPPO was smaller than in admittance control, indicating that the human operator and the robot reached a consensus when performing the board-insertion task.